\documentclass[letter, 10 pt, conference]{ieeeconf}  
\IEEEoverridecommandlockouts                              

\usepackage{hyperref}
\usepackage{placeins}
\usepackage{amssymb}
\usepackage{pifont}
\usepackage{amsmath}
\usepackage{bm}
\usepackage{cancel}
\usepackage{units}
\usepackage{multirow}
\usepackage{ulem}

\usepackage{tikz}
\usetikzlibrary{external,arrows,shapes,automata,backgrounds,petri,fit,arrows.meta,positioning,calc,matrix,chains,scopes}
\usepackage[skins]{tcolorbox}
\usepackage{pgfplotstable}
\usepackage{pgfplots}
\usepackage{subcaption}
\usepackage{algorithmic}
\usepackage{ifthen}
\usepackage{algorithm}
\usepackage{tikz}

\usetikzlibrary{patterns}

\DeclareRobustCommand{\legendsquare}[1]{%
  \tikz[baseline=(a.south)]{\node[#1, inner sep=.8ex, outer sep=0] (a) {};}%
}

\title{\LARGE \bf
Traffic Scenario Clustering by Iterative Optimisation of Self-Supervised Networks Using a Random Forest Activation Pattern Similarity
}

\newcommand{\timeCricitical}{t_\text{0}}

\newcommand{\Occupancy}{\bm{G}_{t}}

\newcommand\copyrighttext{\footnotesize \textcopyright 2021 IEEE. Personal use of this material is permitted. Permission from IEEE must be obtained for all other uses, in any current or future media, including reprinting/republishing this material for advertising or promotional purposes, creating new collective works, for resale or redistribution to servers or lists, or reuse of any copyrighted component of this work in other works.
DOI:\href{tba}{tba}}
\newcommand\copyrightnotice{%
\begin{tikzpicture}[remember picture,overlay]
\node[anchor=south,yshift=10pt] at (current page.south) {\fbox{\parbox{\dimexpr\textwidth-\fboxsep-\fboxrule\relax}{\copyrighttext}}};
\end{tikzpicture}%
}

\author{Lakshman Balasubramanian$^{1}$, Jonas Wurst$^{1}$, Michael Botsch$^{1}$ and Ke Deng$^{2}$
	\thanks{$^{1}$Technische Hochschule Ingolstadt, Research Center CARISSMA, Esplanade
		10, 85049 Ingolstadt, Germany, {\tt\small \{firstname.lastname\}@thi.de}}%
	\thanks{$^{2}$Royal Melbourne Institute of Technology, Melbourne, Australia, 
		{\tt\small \{firstname.lastname\}@rmit.edu.au}}%
}

\begin{document}
\bstctlcite{IEEEexample:BSTcontrol} 

\maketitle
\copyrightnotice 

\thispagestyle{empty}
\pagestyle{empty}

\begin{abstract}

Traffic scenario categorisation is an essential component of automated driving, for e.\,g., in motion planning algorithms and their validation. Finding new relevant scenarios  without handcrafted steps reduce the required resources for the development of autonomous driving dramatically. In this work, a method  is proposed to address this challenge by introducing a clustering technique based on a novel data-adaptive similarity measure, called Random Forest Activation Pattern (RFAP) similarity. The RFAP similarity is generated using a tree encoding scheme in a Random Forest algorithm. The clustering method proposed in this work takes into account that there are labelled scenarios available and the information from the labelled scenarios can help to guide the clustering of unlabelled scenarios. It consists of three steps. First, a self-supervised Convolutional Neural Network~(CNN) is trained on all available traffic scenarios using a defined self-supervised objective. Second, the CNN is fine-tuned for classification of the labelled scenarios.  Third, using the labelled and unlabelled scenarios an iterative optimisation procedure is performed for clustering. In the third step at each epoch of the iterative optimisation, the CNN is used as a feature generator for an unsupervised Random Forest. The trained forest, in turn, provides the  RFAP similarity to adapt iteratively the feature generation process implemented by the CNN.  Extensive experiments and ablation studies have been done on the highD dataset. The proposed method shows superior performance compared to baseline clustering techniques. 


\end{abstract}

\section{Introduction}
\label{Sec:Intro}
In recent years, there have been rapid developments in the field of autonomous driving and driver assistance systems~\cite{Winner2015}. As new and improved autonomous driving functions are introduced, the autonomous system must be capable of handling various driving scenarios. Traffic scenario categorisation is a key component for downstream tasks like path planning~\cite{Chaulwar2017}, behaviour planning and  scenario-based validation methods~\cite{Kruber2018,Kerber2020}. Hence, representative traffic scenarios (e.g. overtake, cut-in, etc.) are necessary for developing and testing the behaviour of autonomous vehicles~\cite{Menzel2018a}. The representative scenarios can be defined by means of  expert knowledge, can be generated from simulations or can be identified from real-world driving data.
\begin{figure}[h!]
	\centering
	\includegraphics[width=\columnwidth]{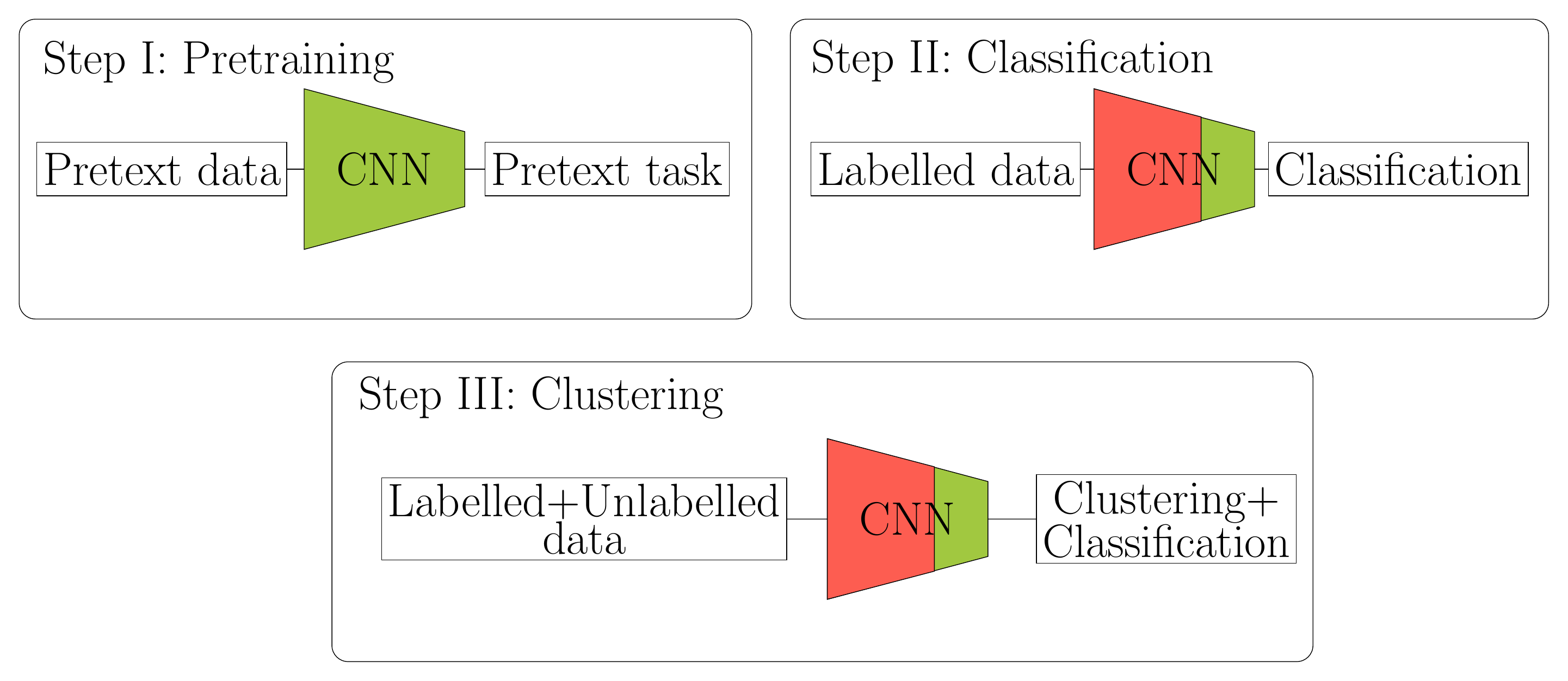}
	\caption{Overview of the three step clustering process. The red portion in the figure shows frozen layers of the CNN and the green portion represents the layers that are trained.}
	\label{fig:overall}
\end{figure}

The approaches using expert knowledge and simulations have important constraints that limit the generation of a list of representative scenarios, such as the limited knowledge of the experts and the  ability of the simulation environment to model the complex interactions between traffic participants. Thus, a promising way to obtain scenario categories is to identify them automatically from real-driving data~\cite{Kruber2018,Harmening2020,Hauer2020}. In this work, the problem of identifying new traffic scenario categories from real-driving data is studied.  Unsupervised scenario clustering methods have been proposed in previous works. In~\cite{Kruber2018,Kruber2019URF}, an Unsupervised Random Forest~(URF) algorithm is proposed to cluster traffic scenarios. The clustering is based on custom features that are selected based on expert knowledge. In~\cite{Hauer2020}, an automated way for deriving traffic scenarios is presented. Even though the dependence on handcrafted features are reduced, still some features and distance measures have to be selected by a human. 
 
In comparison to the aforementioned methods, this work does not rely on any handcrafted features and the proposed method can be applied to all kinds of scenarios irrespective of the number of vehicles present.  More importantly, unlike the works mentioned above, instead of considering this problem as completely unsupervised, this work explores how the knowledge from the available labelled scenarios can help in guiding the clustering of the unlabelled scenarios. This idea is based on the assumption that representations learned for the labelled scenarios can also provide a good representation for the unlabelled scenarios. So, in this work it is assumed that both labelled and unlabelled traffic scenarios are available. 

The clustering process consists of three steps as shown in Fig.~\ref{fig:overall}. The steps are as follows:
\begin{enumerate}
    \item Model initialisation using a self-supervised objective - \textit{pre-training}.
\item Fine tuning the model with labelled scenarios - \textit{classification}.
\item Optimising the model for clustering - \textit{clustering}.
\end{enumerate}

The first step is to pre-train a 3D CNN~\cite{Ji2010} using self-supervision objective i.e., a \textit{pretext} task. The pretext task here means defining a supervised task to train the network without the need for actual ground truth labels, i.\,e., the network is trained using labels that are generated in an automated fashion without any human input. In computer vision, predicting the angle of rotation, colourising images are used as pretext tasks. Hence, both the labelled and unlabelled scenarios are used to pre-train the 3D CNN.  It is important to note that the actual ground truth labels from the labelled scenarios are not used in this step. 

As a second step, a classification head is added to the pre-trained 3D CNN. Only the last layers of the 3D CNN and the classification head  are tuned on the labelled data. This ensures the scenario classification along with  maintaining the general feature extraction capability from the  pre-trained 3D CNN. This is shown in Fig.~\ref{fig:overall}, Step II. The red portion in the figure represents the frozen layers and the green portion represents the fine-tuned layers of the 3D CNN. 

As a third and final step,  a clustering head is added to the fine-tuned 3D CNN. The last layers of the 3D CNN along with the classification and clustering heads are iteratively optimised using both the labelled and unlabelled scenarios. The classification head is optimised using the ground truth labels from the labelled scenarios. The clustering head is optimised using a data-adaptive similarity generated from an URF algorithm called Random Forest Activation Pattern (RFAP) similarity. The implementation of the architecture is made publicly available\footnote{https://github.com/lab176344/TrafficScenarios-RFAPsClustering}.

The contributions of this work are the following:
\begin{itemize}
	\item Introduction of a self-supervised pretext task for traffic scenarios.
	\item Introduction of novel data-adaptive features called RFAPs. The RFAPs provide a data-adaptive similarity measure for the unlabelled data.
	\item Presenting a method for unsupervised clustering of unlabelled traffic scenarios given few labelled scenarios. 
	\item Comprehensive analysis and ablation studies on the proposed methods using the highD~\cite{Krajewski2018a} dataset.
\end{itemize}

The remainder of the paper is organised as follows: Section~\ref{Sec:Rel} presents the related work. Section~\ref{Sec:Meth} discusses the proposed method. Section~\ref{Sec:Exp} illustrates the experiments and analysis on the highD dataset. Section~\ref{Sec:Abla} discusses the ablation studies. Finally, the paper is concluded in Section~\ref{Sec:Con}.

\section{Related Works}
\label{Sec:Rel}
\subsection{Traffic Scenario Clustering}
Unsupervised traffic scenario clustering methods have been studied in previous works~\cite{Kruber2018,Harmening2020,Hauer2020,}. There are also methods~\cite{Demetriou2020,Langner2019} that focuses on trajectory clustering which compares and clusters trajectories from a single vehicle. Since this work focus on clustering scenarios with multiple traffic participants, publications about clustering trajectories are not discussed further. In~\cite{Kruber2019URF}, an URF algorithm is proposed to identify clusters from simulated driving data. The authors suggest using the path based proximity from an URF algorithm along with Hierarchical Clustering~(HC) to solve the task. But, the method relies on features selected by experts and the clusters are selected visually, which limits the amount of data that can be clustered at a single time. In~\cite{Hauer2020}, an automatic scenario clustering method on real-world driving data is proposed. The automatic clustering uses dynamic time warping to compare distances between trajectories based on handcrafted features and the  generated distance measure is used to construct scenario clusters. In~\cite{Kerber2020}, a spatial filter  to determine the relevant target objects  around the ego and a custom distance metric along with HC is proposed to cluster scenarios. 

There are also approaches like~\cite{Wang2018a} which cluster traffic scenarios based on the interaction between two vehicles using Long Short Term Memory (LSTM)+CNN. A deep learning based method for clustering traffic scenarios  is presented in~\cite{Harmening2020}. Spatio-temporal autoencoders and recurrent neural networks are used for solving the task of traffic scenario clustering.

In comparison to the above mentioned works except~\cite{Harmening2020}, which does feature engineering, the scenarios in this work are described as a sequence of occupancy grids. Such a representation can be generated from most of the common autonomous vehicle sensor suites. Also, in this work the scenarios considered are not limited by the number of traffic participants around the ego. As opposed to all the clustering methods discussed above, the problem setting addressed in the proposed method is different, this work utilises labelled traffic scenarios in guiding the clustering of unlabelled traffic scenarios. 

\subsection{Method Comparison}
The method presented in this work is based on~\cite{Han2020}, but extends the architecture in the following ways: (1). The traffic scenarios used in this work are described as time series data as opposed to images in~\cite{Han2020}, (2). A new self-supervised objective is introduced for training a 3D CNN for traffic scenarios, (3). A data-adaptive similarity measure based on novel features generated from a RF algorithm~\cite{Breiman2001} is introduced for clustering purposes. Also, intended application of the proposed method is different. 


\begin{figure*}[h!]
    \vspace{2mm} 
	\centering
	\includegraphics[width=0.85\linewidth]{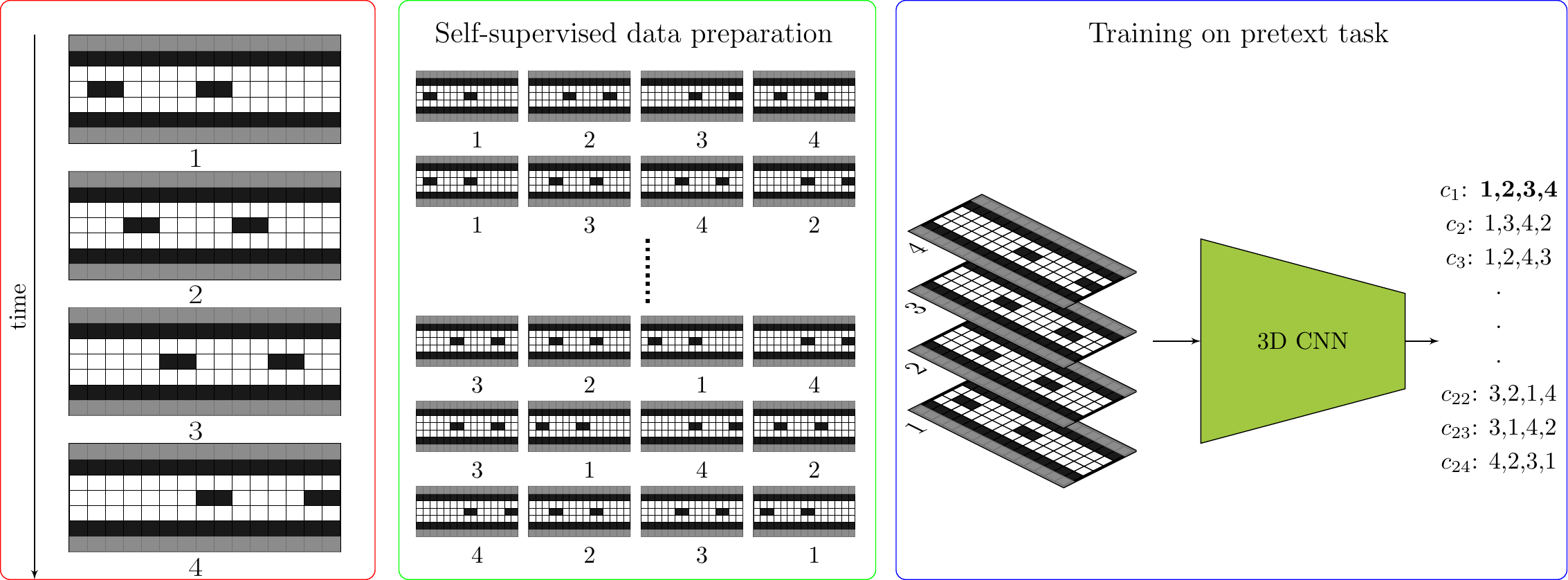}
	\caption{Self-supervised learning for traffic scenarios. The red box shows the original scenario in the correct temporal order, the green box shows the self-supervised data preparation, and the blue box shows the pretext task of classifying the input to one of the $24$ classes.}
	\label{Fig:ssl_traffic}
\end{figure*}
\section{Methodology}
\label{Sec:Meth}
This section details the clustering methodology. A dataset with unlabelled traffic scenarios  $\mathcal{D}_\mathrm{u}=\left\{\textbf{G}_1^\mathrm{u},\ldots,\textbf{G}_{M_\mathrm{u}}^\mathrm{u}\right\}$, where the traffic scenarios are represented as a sequence of occupancy grids $\textbf{G}$, is available. Also, a dataset with labelled traffic scenarios  $\mathcal{D}_\mathrm{l}=\left\{(\textbf{G}_1^\mathrm{l},y_1),\ldots,(\textbf{G}_{M_\mathrm{l}}^\mathrm{l},y_{M_\mathrm{l}})\right\}$, where $y_{m_\mathrm{l}}$ is the scenario label. 
$M_\mathrm{l}$ and $M_\mathrm{u}$ are the number of labelled and unlabelled data respectively. The clustering is realised in a three step process. In step I, a 3D CNN is pre-trained on a defined pretext task using both $\mathcal{D}_\mathrm{l}$ and $\mathcal{D}_\mathrm{u}$ to learn robust feature representations (see Sec.~\ref{SubSec:SSL}). In step II, a classification head is added to the pre-trained 3D CNN. The classification head and the last layers of the 3D CNN are fine-tuned only using $\mathcal{D}_\mathrm{l}$ (see Sec.~\ref{SubSec:StepII}. In step III, a clustering head is added to the fine tuned 3D CNN. An iterative optimisation procedure is performed on the last layers of the 3D CNN, the clustering and the classification heads. For solving the clustering task using both $\mathcal{D}_\mathrm{l}$ and $\mathcal{D}_\mathrm{u}$ are used (see Sec.~\ref{SubSec:StepIII}). The traffic scenario representation used in this work is described first before explaining the methodology in detail.

\subsection{Traffic Scenario Representation}
\label{Sec:TScenarios}
Traffic scenarios in this work are described as introduced in~\cite{Chaulwar2017}, a discretised space-time representation of the environment around the ego,  from the time $t_{-3}$ to $t_{0}$. The $t_{-3}$ refers to a time before $t_{0}$, i.\,e., a time before the traffic situation becomes interesting. The traffic scenario at each time instance is represented as a 2D occupancy grid  $\Occupancy\in\mathbb{R}^{I \times J}$. The occupancy probability of each cell $\Occupancy(i,j)$ is assigned either as 1 for occupied space, 0 for free space or 0.5 for an unknown region. For a time span of $t_{0}-t_{-3}=1.5$s and $\Delta t$ of $0.5$s, a traffic scenario is represented as $\textbf{G}=\left[\bm{G}_{t_{-3}},\bm{G}_{t_{-2}},\bm{G}_{t_{-1}},\bm{G}_{t_{0}} \right]$, where $\textbf{G}\in\mathbb{R}^{I\times J \times N_{\text{ts}}}$ with $I$ rows, $J$ columns and a depth of $N_{\text{ts}}$. The number of time steps is $N_{\text{ts}}=1+\frac{t_{0}-t_{-3}}{\Delta t}$. An exemplary  representation is shown in Fig.~\ref{Fig:ssl_traffic}. The  trigger used to determine  the time  $\timeCricitical$  at which the traffic situation becomes interesting is determined based on Time-Headway~(THW).

\subsection{Step I: Self-Supervised Learning for Traffic Scenarios}
\label{SubSec:SSL}
The main aim in this step is to train a 3D CNN network with a pretext task to provide robust low-level features for the traffic scenarios. To do this, the most naive way is to train the 3D CNN only on the labelled data first. The trained model can be used as a feature extractor for the unlabelled data and clustering is done on the extracted features. But, this does not guarantee that a model learned from the labelled data will generate good features for the unlabelled data.  The model might be biased towards the classification of the labelled data.  Hence, this work proposes a self-supervised pre-training similar to~\cite{Han2020,Chen2020} to generate robust low-level features which later can be fine tuned for the clustering. 

 The self-supervised objective does not require any data annotations, so it can be applied for both labelled and unlabelled data. The idea behind self-supervision is that in the course of solving the self-supervised objective the model will learn some semantic structure in the data and learn robust low-level features. The self-supervised objective used in this work is a pretext task.

The pretext task is to predict the correct temporal order given a sequence of occupancy grids. Consider a sequence of four occupancy grids $\textbf{G}$  that describe a traffic scenario. The sequence of four occupancy grids in the correct temporal order can be shuffled and $24$ different combinations of sequences can be produced i.e., say the correct temporal order is $(1,2,3,4)$ after shuffling one can have $24$ different orders like $\{c_\mathrm{1}=(1,2,3,4),\ldots,c_\mathrm{23}=(3,1,4,2),c_\mathrm{24}=(4,2,3,1)$\}. This can be seen in the green box of Fig.~\ref{Fig:ssl_traffic}. So, each of the shuffled $24$ sequence of grids can be assigned to one of the order from $\{c_\mathrm{1},\ldots,c_\mathrm{24}\}$. The problem this way is converted to a $24-$class supervised classification task for assigning an input grid $\textbf{G}$ to one of the $24$ orders $\{c_\mathrm{1},\ldots,c_\mathrm{24}\}$. This can be seen in the blue box of Fig.~\ref{Fig:ssl_traffic}, where $\textbf{G}$ with the order $(1,2,3,4)$ is given as input and the network chooses $c_\mathrm{1}$ as the output. The labels for this pretext task is generated without using any ground truth labels from $\mathcal{D}_\mathrm{l}$  are used here.   Temporal order shuffling has been explored as pretext task for video classification in~\cite{Ishan2016}. The rationale behind temporal shuffling is to make the model understand the temporal structure in the data by reasoning out how the vehicles moves in the scenario snippets. The classification here in this work is done with a 3D CNN $f(\textbf{G})$, where the convolution happens both in spatial and temporal dimensions. The 3D CNN is trained with categorical cross entropy to classify a shuffled $f(\textbf{G})$ to one of the $24$ classes as shown in the blue box in Fig.~\ref{Fig:ssl_traffic}. 
 \begin{figure*}[h!]
	\centering
	\includegraphics[width=0.8\linewidth]{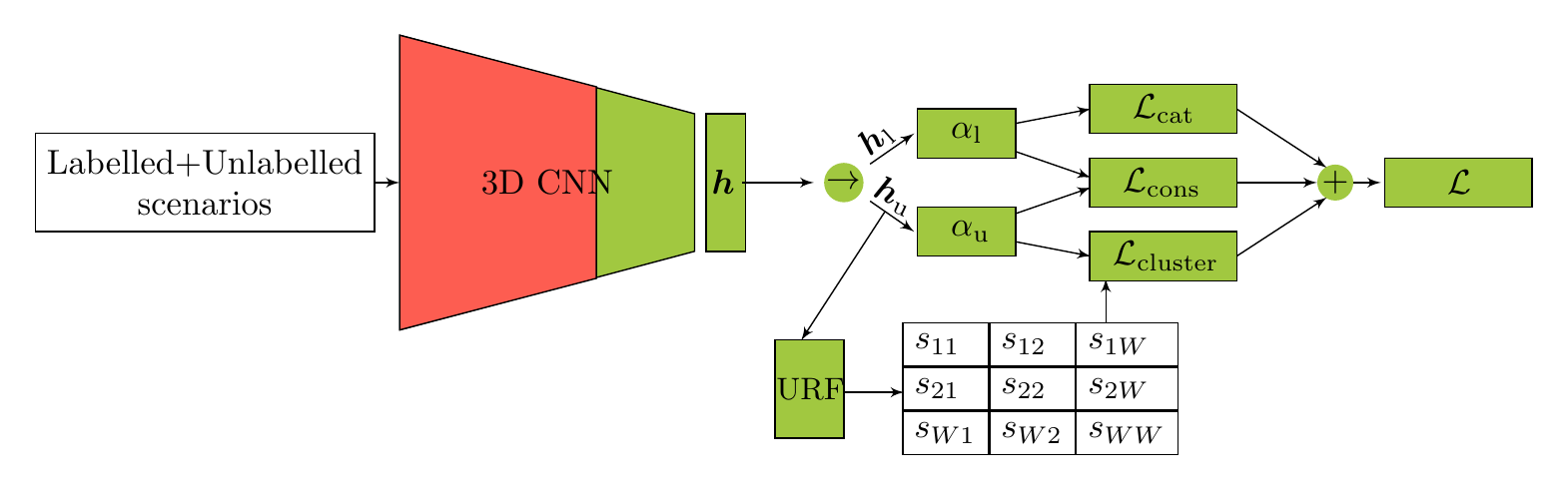}
	\caption{Iterative optimisation procedure - single epoch. The operator $\rightarrow$ separates the labelled and unlabelled data.}
	\label{fig:RFSimProp}
\end{figure*}
\subsection{Step II: Fine-Tuning with Labelled Data}
\label{SubSec:StepII}

The goal in this step is to fine-tune $f(\textbf{G})$ which is pre-trained in step I using the ground truth labels from $\mathcal{D}_\mathrm{l}$.  The mapping $\textbf{G}\mapsto\bm{y}$ is realised, where $\bm{y}\in  \mathbb{R}^{K}$  is the one-hot representation of the scenario label with $K$ labelled classes. Before fine-tuning, the fully connected layers from $f(\textbf{G})$, which were trained for the $24$-class problem, are removed. A new classification head $\alpha_\mathrm{l}$ is added to $f(\textbf{G})$ to classify the input to one of $K$ classes.  The classification head here refers to a fully connected layer followed by a softmax activation layer. Only the last layers of $f(\textbf{G})$ and $\alpha_\mathrm{l}$ are tuned to retain the robust low-level features from pre-training and to learn only high level features for scenario classification. The network $f(\textbf{G}): \mathbb{R}^{I\times J \times N_{\text{ts}}} \to \mathbb{R}^F$ takes a traffic scenario $\textbf{G}\in\mathbb{R}^{I\times J \times N_{\text{ts}}}$ as input and produces the vectorised representation $\bm{h}\in\mathbb{R}^F$, where $F$ is the dimensionality of the vectorised representation. Given $\mathcal{D}_\mathrm{l}$, the classification head $\alpha_\mathrm{l}(\bm{h}^\mathrm{l}): \mathbb{R}^F\to\mathbb{R}^{K}$ takes the representation $\bm{h}^\mathrm{l}$ as input and produces the target vector $\bm{\hat{y}} \in  \mathbb{R}^{K}$, where $K$ is the number of classes in the labelled data. The last layers of the network $f(\textbf{G})$ and the classification head $\alpha_\mathrm{l}(\bm{h}^\mathrm{l})$ are trained using the categorical cross entropy, 

\begin{equation}
	\mathcal{L}_{\mathrm{cat}} = -\frac{1}{M}\sum_{m=1}^{M} \sum_{k=1}^{K} \bm{y}_{m,k} \mathrm{log}(\bm{\hat{y}}_{m,k}).
\end{equation}

\subsection{Step III: Iterative Optimisation with Labelled and Unlabelled Data}
\label{SubSec:StepIII}
The goal of  clustering is to divide the given unlabelled data into $Q$ number of groups. To realise this as a final step $f(\textbf{G})$ is  iteratively optimised  to learn feature representations for the unlabelled data in $\mathcal{D}_\mathrm{u}$. In this section the following contents are discussed: the iterative optimisation procedure to learn feature representations for the scenarios in $\mathcal{D}_\mathrm{u}$, and similarity calculation based on RFAPs. 
\subsubsection{Iterative Optimisation}
The intent of doing iterative optimisation task is to fine-tune the last layers and the heads of $f(\textbf{G})$ using the losses $\mathcal{L}_\mathrm{cluster}$, $\mathcal{L}_\mathrm{cat}$ and $\mathcal{L}_\mathrm{cons}$ to cluster the unlabelled scenarios. $\mathcal{L}_\mathrm{cluster}$ is the clustering loss to learn feature representations for unlabelled scenarios. $\mathcal{L}_\mathrm{cat}$ is the categorical cross entropy loss which retains the knowledge from labelled scenarios. $\mathcal{L}_\mathrm{cons}$ is used to ensure stability when training with unlabelled scenarios.

The clustering is realised by iteratively-tuning $f(\textbf{G})$ trained on both $\mathcal{D}_\mathrm{u}$ and $\mathcal{D}_\mathrm{l}$, i.\,e. tuning the network epoch by epoch.  In a single epoch the following procedures are performed. First, the representation set $\mathcal{D}_\mathrm{u}^\mathrm{h}=\left\{\bm{h}_1^\mathrm{u},\bm{h}_2^\mathrm{u},\ldots,\bm{h}_N^\mathrm{u}\right\}$ for the scenarios in $\mathcal{D}_\mathrm{u}$ is extracted. Followed by that, an URF algorithm $g(\bm{h}^\mathrm{u})$  is trained on the extracted features $\mathcal{D}_\mathrm{u}^\mathrm{h}$. Given a mini-batch, the trained URF algorithm can then be used to compute the proximity/similarity matrix $\bm{S}$. The similarity in this work is based on RFAPs which is explained in Section~\ref{Sec:RFAP}. Using  $\bm{S}$ and extracted features $\bm{h}_\mathrm{u}$ and $\bm{h}_\mathrm{l}$ of the labelled and unlabelled scenarios, the three losses mentioned above are constructed and used for optimisation in a single epoch. These procedures are repeated for a selected number of epochs. At each epoch a new URF model $g(\bm{h}^\mathrm{u})$ is trained using the extracted features $\mathcal{D}_\mathrm{u}^\mathrm{h}$ at that epoch. The complete schematic procedure for a single epoch is shown in Fig.~\ref{fig:RFSimProp}. The procedures at a single epoch are discussed in detail.

\paragraph{Clustering Loss}
Given an URF  $g(\bm{h}^\mathrm{u})$ trained on  $\mathcal{D}_\mathrm{u}^\mathrm{h}$, the similarity matrix $\bm{S}$ is  computed for a mini-batch of unlabelled traffic scenarios using the RFAPs. The computation of the similarity with RFAPs is explained in detail in Section~\ref{Sec:RFAP}. The value at $S_{ij}\in(0,1]$, gives the pairwise-similarity between the $i$-th and $j$-th scenario from the mini-batch denoting how similar the scenarios are.  The optimisation objective for clustering is constructed using the matrix $\bm{S}$. A new clustering head  $\alpha_\mathrm{u}(\bm{h}^\mathrm{u}): \mathbb{R}^F \to \mathbb{R}^{Q}$ is added to $f(\textbf{G})$ parallel to the head $\alpha_\mathrm{l}(\bm{h}^\mathrm{l})$. Here, $Q$ is the number of clusters in the dataset $\mathcal{D}_\mathrm{u}$. Following~\cite{Rebuffi2020,Wada2019,Guo2017}, the parameter $Q$ is assumed to be known but in Section~\ref{Sec:Abla} a study is performed, where the parameter $Q$ is kept as unknown and $Q$ is estimated as suggested in~\cite{Han2019}. The clustering loss to fine-tune the $f(\textbf{G})$ for the unlabelled dataset $\mathcal{D}_\mathrm{u}$ is based on binary cross entropy. It is given by

\begin{multline}
	\mathcal{L}_\mathrm{cluster} = \frac{1}{W^2}\sum_{i=1}^{W}\sum_{j=1}^{W} {S}_{ij}~\mathrm{log}~\mathrm{P}(i=j)+\\(1-{S}_{ij})~\mathrm{log}~ \mathrm{P}(i\neq
	j),
\end{multline}
 where $W$ is the mini batch size. $\mathrm{P}(i=j)$ denotes the probability that the traffic scenario $i$ and $j$ are in the same cluster. A similar loss was used in~\cite{Han2020,Rebuffi2020}, but $\bm{S}$ in this work is given by the data-adaptive similarity from the RFAPs. As shown in~\cite{Rebuffi2020}, if the number of clusters $Q$ is fixed and $i,j$ are independent, $\mathrm{P}(i=j)$ can be modelled as the inner product between the vectors $\alpha_\mathrm{u}(\bm{h}_i^\mathrm{u})$ and $\alpha_\mathrm{u}(\bm{h}_j^\mathrm{u})$. The final clustering loss is given as

\begin{multline}
	\mathcal{L}_\mathrm{cluster} = \frac{1}{W^2}\sum_{i=1}^{W}\sum_{j=1}^{W} {S}_{ij}\mathrm{log}( \alpha_\mathrm{u}(\bm{h}_i^\mathrm{u})^\top \alpha_\mathrm{u}(\bm{h}_j^\mathrm{u}))+ \\(1-{S}_{ij})\mathrm{log}(1-( \alpha_\mathrm{u}(\bm{h}_i^\mathrm{u})^\top \alpha_\mathrm{u}(\bm{h}_j^\mathrm{u}))).
\end{multline}

Now as the clustering loss is defined, the last layers of $f(\textbf{G})$ and the new clustering head can be fine-tuned for the clustering task. An argmax on the vector $\alpha_\mathrm{u}(\bm{h}^\mathrm{u})$  will provide the cluster number to which an input scenario belongs. 

But, training only on the unlabelled data will destroy the representation learned for the labelled dataset, leading to catastrophic forgetting~\cite{Rebuffi2016}  . Hence, both heads $\alpha_\mathrm{l}$ and $\alpha_\mathrm{u}$ are trained in parallel on categorically cross entropy using the dataset $\mathcal{D}_\mathrm{l}$ and the clustering loss using the dataset $\mathcal{D}_\mathrm{u}$ respectively. Additionally, the head $\alpha_\mathrm{l}$ is extended to classify the clusters identified by the clustering head  $\alpha_\mathrm{u}$. So, the head $\alpha_\mathrm{l}$ does the mapping $\mathbb{R}^F \to \mathbb{R}^{K+Q}$ and it is trained on the categorical cross entropy. 
This way the knowledge of the labelled traffic scenarios is also preserved when optimising for clustering the unlabelled traffic scenarios. The knowledge from the labelled traffic scenarios is shown to improve clustering accuracy from in Section~\ref{Sec:Abla}.

\paragraph{Consistency Loss}
The similarity ${S}_{ij}$ between two given traffic scenarios is updated every epoch as the representation is optimised by the categorical cross entropy and the clustering losses. This might lead to instability or inconsistency in the training for the unlabelled dataset. Following~\cite{Han2020,Rebuffi2020}, a consistency constrain, which is used in semi-supervised learning setting~\cite{Miyato2019}, is also introduced. The consistency loss constrains the model to produce the same output for a given traffic scenario $\textbf{G}_i \mapsto \bm{h}_i$ and an augmented version of the traffic scenario $\hat{\textbf{G}}_i \mapsto \bm{\hat{h}}_i$ (e.g. random erasing, adding random noise). The consistency loss is given by

\begin{multline}
	\mathcal{L}_\mathrm{cons} = \frac{1}{M_\mathrm{l}}\sum_{i=1}^{M_\mathrm{l}}( \alpha_\mathrm{l}(\bm{h}_i^\mathrm{l})-\alpha_\mathrm{l}(\bm{\hat{h}}_i^\mathrm{l}))+\\	\frac{1}{M_\mathrm{u}}\sum_{i=1}^{M_\mathrm{u}}( \alpha_\mathrm{u}(\bm{h}_i^\mathrm{u})-\alpha_u(\bm{\hat{h}}_i^\mathrm{u})).
\end{multline}

\paragraph{Total Loss}
The total loss used to optimise $f(\textbf{G})$ for the clustering task while keeping the knowledge from the labelled data is given by,
\begin{equation}
	\mathcal{L} = \mathcal{L}_\mathrm{cat} + \mathcal{L}_\mathrm{cluster} + \omega(\beta)\mathcal{L}_\mathrm{cons}.
\end{equation}

Following~\cite{Tarvainen2017}, $\omega(\beta)=\lambda \mathrm{exp}(-5(1-\frac{\beta}{T})^2)$ is the ramp-up function where $\beta$ being the epoch, $T$ is the ramp-up length and $\lambda\in\mathbb{R}_+ $. Using $\mathcal{L}$, the last layers of the $f(\textbf{G})$ and the heads $\alpha_\mathrm{l}$ and $\alpha_\mathrm{u}$ are tuned. This procedure is repeated a given number of epochs
. 
\subsubsection{Data-adaptive similarity with RFAPs}
\label{Sec:RFAP}
In this work, the similarity matrix $\bm{S}$ is defined by a data-adaptive similarity measure based on features called RFAPs. In the following, the generation of RFAPs from a URF algorithm and calculation of  $\bm{S}$ with RFAPs are explained. 
\paragraph{Generation of RFAPs}
The RFAPs are generated by an URF algorithm $g(\bm{h}^\mathrm{u})$ trained on the extracted features $\bm{h}^\mathrm{u}$ of the scenarios from $\mathcal{D}_\mathrm{u}$. The generation of RFAPs and the calculation of similarity will be discussed in this section. 

The generation of RFAPs is based on a novel encoding scheme for indexing the nodes of the URF trees. The RFAPs in turn will be used to calculate the data-adaptive similarity. Let the $B$ trees in the URF~$\{T_1,\ldots,T_B\}$  be fully grown. The $b$-th tree $T_b$ divides the input space $\mathbb{R}^F$ into many small hypercubes. So, if two data points $\bm{h}_{i}$ and $\bm{h}_{j}$ end up in the same hypercube they are similar to each other. This is termed as proximity in~\cite{Breiman2001}. Similarly, the paths taken by the data points to reach the terminal nodes  can also provide a similarity measure by comparing the common paths taken~\cite{Kruber2019URF}. 

The paths taken by the data point~$\bm{h}_i$ in all the trees can be represented by the vector $\bm{r}_i\in\mathbb{N}^B$. Each element of $\bm{r}_i$ is an ordinal number representing a path through a tree. By using a special node encoding, a single number is sufficient to identify a specific path in a tree. In the following, the encoding of a data point's path  in one tree is presented. This encoding is the basis for RFAPs. In a first step, all nodes of each tree have to  be indexed. Looking at one specific node in the tree $T_b$ the following quantities are used for indexing: the maximum depth of the tree $d_{b}$, the depth of the considered node $k_{b}\in\{1,\ldots,d_{b}\}$, and the number of nodes $N_b$ in $T_b$.

\begin{algorithm}
	\caption{Indexing for a tree in the RF}
	\begin{algorithmic}[1]
		\renewcommand{\algorithmicrequire}{\textbf{Input: }}
		\renewcommand{\algorithmicensure}{\textbf{Output:}}
		\REQUIRE $T_b$, $d_{b}$, $N_b$
		\ENSURE  Indexed  $T_b$
		\STATE $id ^1 \gets 0$

		\FOR {$n = 2$ to $N_b$}	
		\STATE 	$id^\mathrm{pr}$ = {getParentNodeid($id ^n$)}
		\IF {n.isleft}
		\STATE $id ^n \gets id^\mathrm{pr}+10^{d_{b}-k_{b}}$
		\ELSE
		\STATE $id^n \gets id^\mathrm{pr}+2*10^{d_{b}-k_{b}}$	
		\ENDIF
		\ENDFOR
	\end{algorithmic}
	\label{AlgorthimIndex}
\end{algorithm}

\begin{figure}[h!]
    \vspace{2mm} 
	\centering
	\includegraphics[width=0.8\columnwidth]{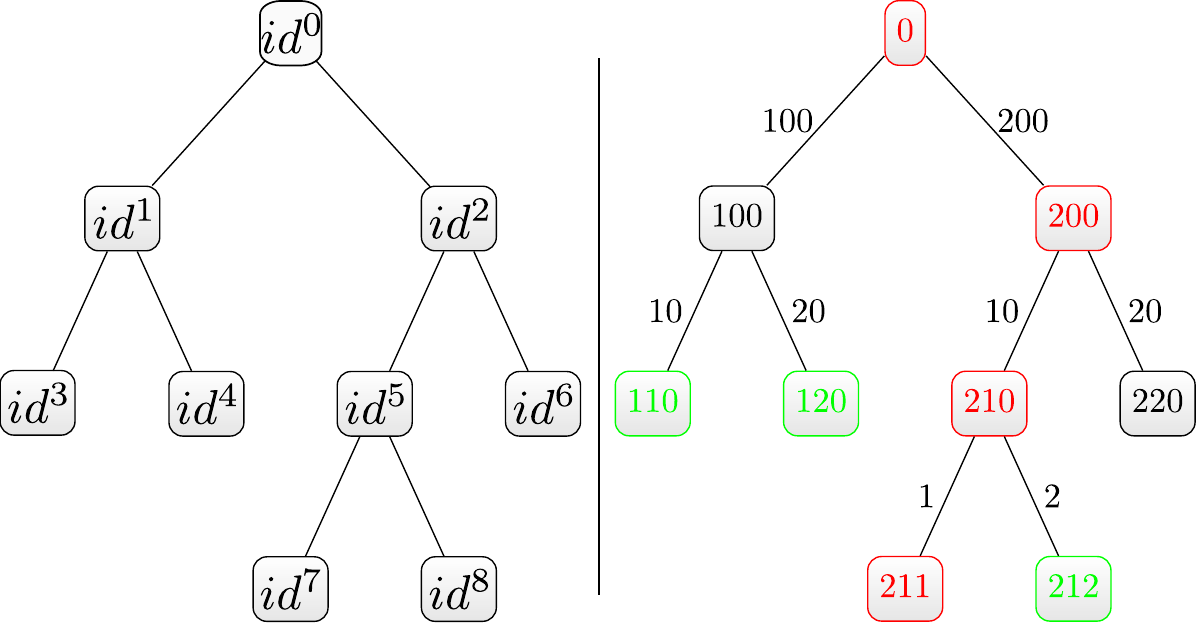}
	\caption{RF tree before indexing (left) and indexed tree (right).}
	\label{fig:indexedtree}
\end{figure}

The algorithm for indexing a tree is depicted in Alg.~\ref{AlgorthimIndex}. The index of the root node $id^1$ at $k_{b}=1$ in a tree is set to 0. Each level in a tree is indexed as follows: if the node is a left child node then the index for the child node is assigned as the sum of the index of its parent node $id^\mathrm{pr}$ and $10^{d_{b}-{k_{b}}}$, while the right child node is indexed as the sum of its $id^\mathrm{pr}$ and $2*10^{d_{b}-{k_{b}}}$.  With this indexing of nodes, the path of  data point $\bm{h}_i$ in the tree $T_b$ is encoded in the ordinal number assigned to the terminal node reached by $\bm{h}_i$.  An example indexed tree $T_b$ is shown in Fig.~\ref{fig:indexedtree} with $N_b=9$ and $d_{b}=4$. Assume a data point $\bm{h}_i$, which is reaching the terminal node $211$. The index 211 is encoding the path shown in red: $0 \to 200 \to 210 \to 211$. Considering all $B$ trees in a RF, the RFAP-representation for the data point $\bm{h}_i$ is represented as  $\bm{r}_{i}=\begin{bmatrix}
	id_{1}^i,
	id_{2}^i,
	\dots,
	id_{B}^i
\end{bmatrix}^T$, where $id_{b}^i$ denotes the index of the terminal node reached by $\bm{h}_i$ in the tree $T_b$.

\paragraph{Calculating Similarity with RFAPs} 
The advantage of RFAPs is that the complete path taken by two data points can be used to calculate similarity. The digits of the RFAP indices encodes the complete path information, The more similar the digits are  between two RFAP indices, the deeper the paths are shared to reach the terminal nodes. Hence, the similarity with RFAPs is calculated using the hamming distance. The indices of the nodes are of equal length, and if the indices are treated as sequences of digits, the similarity ${S}_{ij}$ can be given as,
\begin{equation}
{S}_{ij} =1- \frac{1}{B}\sum_{b=1}^{B} \frac{\vert\left\lbrace o\in\left\lbrace 1,\dots,\vert\bm{r}_{i}^b\vert\right\rbrace\vert\bm{r}_{i}^b\left[o\right]\not=\bm{r}_{j}^b\left[o\right])\right\rbrace\vert}{\vert\bm{r}_{j}^b\vert}.
\end{equation}.

Here, $\bm{r}_{i}^b$ refers to the RFAP index at the $b$-th position in $\bm{r}_{i}$. The numerator checks how many digits are not equal when comparing  $\bm{r}_{i}$ to $\bm{r}_{j}$. The $\vert.\vert$ returns the digits length of elements in the vector $\bm{r}_{i}^b$. For e.\,g., in  Fig.~\ref{fig:indexedtree}, with the terminal node indices $211$ and $212$, the similarity in a single tree is $1-(1/\vert211\vert)$, where $\vert211\vert=3$. A ${S}_{ij}$ value of $1$ means the two data points $i$ and $j$ took the same path in all the trees.

In summary, the clustering is done in a three step process: As a first step, in order to get robust low-level features a 3D CNN is trained on a pretext task. As a second step, the  trained 3D CNN is fine-tuned on  labelled classes. As the final step, the fine-tuned network is iteratively optimised using the three losses combined for clustering while preserving the knowledge from the labelled data.

\section{Experiments and Results}
This section reports the experiments and results of the proposed method applied on the highD dataset. 
\label{Sec:Exp}
\subsection{Dataset}

The highD dataset is a naturalistic vehicle trajectory dataset recorded using drones on German motorways. The dataset consists of $16.5$ hours of drone video records in six different locations and contains around $110\,000$ vehicles. The data analysed in this work is restricted to  traffic scenarios where the ego vehicle has a leading vehicle. Since this work focuses on traffic scenarios to validate the clustering method, $7$ common highway scenarios are extracted from the dataset:

\begin{enumerate}\addtocounter{enumi}{0}
    \setcounter{enumi}{0}

	\item Ego lane change to the right lane,
	\item Ego lane change to the left lane,
	\item Leader cutting into ego's lane from the left lane,
	\item Leader lane change from ego's lane to the left lane,
	\item Leader cutting into ego's lane from the right lane,
	\item Ego following the leader in a lane,
	\item Leader lane change from ego's lane to the right lane.	
\end{enumerate}
	
As described in~\ref{Sec:TScenarios}, THW$ < 4$\unit{s} is used as the criterion for finding interesting scenarios. The environment at each time instance is represented with $\Occupancy$ of span $15$\unit{m}~$\times~200$\unit{m} and a resolution of $0.5$\unit{m}$~\times~1$\unit{m}. The interval  $t_\text{lb}-t_0$ and $\Delta t$ used in this work are 2\unit{s} and 0.5\unit{s} respectively. So, a traffic scenario is represented with \textbf{G} of size $30\times200\times4$. The grids are generated in an ego-centric fashion fixed at $t_0$. In total $4480$ scenarios are extracted and is split as $70\%$  for training, $10\%$ for validation and $20\%$ for testing. 

\subsection{Implementation Details}
The $f(\textbf{G})$ is a 3D-Resnet~\cite{Hara2017}, a 3D version of the normal Resnet~\cite{He2016} with spatio-temporal operations.  The URF is trained with $B=500$ trees. The parameters  $T$  and $\lambda$ are set to $100$ and $5$ respectively. The $f(\textbf{G})$ is trained using SGD optimiser~\cite{Sutskever2013} with moment and decay and with a batch size $W$ of $32$. Only the last Residual block of the 3D-ResNet is trained in the Step II and III.

\subsection{Evaluation Metric}
The evaluation metric used for measuring the clustering accuracy (ACC) is the unsupervised clustering accuracy~\cite{Yang}. The best mapping between the labels obtained from clustering and the ground truth is computed by the Hungarian algorithm. The ACC is defined as follows

\begin{equation}
	\mathrm{ACC} = \max\limits_m  \frac{\sum_{j=1}^{M_\mathrm{u}} 1(y_j = m(c_j^\mathrm{u}))}{M_\mathrm{u}},
\end{equation}

where $y_j$ is the ground truth and $c_j^\mathrm{u}$ is the predicted label for the unlabelled sample $\textbf{G}_j$. The range of the ACC is $[0,1]$, with 1 referring perfect clustering. 

\subsection{Baselines}
The proposed method is compared with the following methods: (a) $K$-means clustering~\cite{MacQueen1967} directly on the dataset $\mathcal{D}_\mathrm{u}$. (b) A spatio Temporal Autoencoder + Heirrarchial Clustering (STAE+HC)~\cite{Harmening2020}, a spatio-temporal autoencoder is trained on the dataset  $\mathcal{D}_\mathrm{u}$ and HC is performed on the latent space of the autoencoder.  (c) Autonovel~\cite{Han2020}, a spatio-temporal extension of the proposed method in~\cite{Han2020} is done and the method uses rank statistics as similarity measure for clustering. (d) Comparison with other similarity measures like cosine, L$2$, rank statistics~\cite{Han2020} and the similarity from a K-Nearest Neighbour~(KNN) algorithm for determining $\bm{S}$.   
\subsection{Results}
\subsubsection{Clustering}
In this experiment, the following problem setting is used. The first $4$  classes out of the available $7$ scenarios are treated as labelled classes and the remaining $3$ classes are treated as unknown. In the ablation study, an analysis is also conducted with a different combination of labelled and unlabelled classes. The aim of this experiment is to measure the clustering accuracy on the $3$ unlabelled classes. As suggested in~\cite{Rebuffi2020,Wada2019}, the experiment is repeated $5$ times and an average of the accuracy~(ACC) is reported in the Table~\ref{Tab:highDcluster}. There, it can be seen that the proposed method outperforms the baselines and provides good clustering accuracy.

\begin{table}[h]
	\centering
	\caption{highD - Clustering accuracy.}
	\begin{tabular}{| c | c |} 
		\hline
		Method &  ACC ($\uparrow$)  \\ 
		\hline\hline
		$K$-means~\cite{MacQueen1967} & 0.391 \\  
		STAE+HC~\cite{Harmening2020} & 0.52 \\
		Autonovel~\cite{Han2020} & 0.794 \\
		Proposed method~(RFAPs) & \textbf{0.810} \\
		\hline
	\end{tabular}
	\label{Tab:highDcluster}
\end{table}
\subsubsection{Comparison with Other Similarities}
In this experiment, the proposed RFAP based pairwise similarity used in the mini-batch for clustering is compared with standard similarity measures like cosine, L$2$, rank statistics~\cite{Han2020} and similarity from KNN. From the results shown in Table~\ref{Tab:highDsimilarity}, it can be seen the RFAP  similarity provides superior performance when compared to other standard similarity measures.
\begin{table}[h]
	\centering
	\caption{Clustering accuracy with different similarities.}
	\begin{tabular}{| c | c | c | c | c | c |} 
		\hline
		Similarity &  cosine & L$2$ & KNN &  rank~\cite{Han2020}  &  RFAPs \\ 
		\hline\hline
		ACC ($\uparrow$) & 0.707 & 0.703 & 0.793 & 0.794 & \textbf{0.810} \\  
		\hline
	\end{tabular}
	\label{Tab:highDsimilarity}
\end{table}

\section{Ablation Study}
\label{Sec:Abla}
\subsection{Importance of Step I}
To underline the importance of self-supervised initialisation, another experiment is setup. For this, as a first step all the layers of the 3D CNN is trained on the labelled classes without any Self-Supervised Learning~(SSL). As a second step, the iterative optimisation as described in the Section~\ref{Sec:Meth}. As seen in Table~\ref{Tab:highDSS}, the ACC  improves with SSL initialisation. 
\begin{table}[h]
	\centering
	\caption{ACC with and without SSL initialization.}
	\begin{tabular}{| c | c |} 
		\hline
		Method &  ACC ($\uparrow$)  \\ 
		\hline\hline
		with SSL & \textbf{0.810} \\  
		without SSL & 0.537 \\
		\hline
	\end{tabular}
	\label{Tab:highDSS}
\end{table}

\subsection{Estimating $Q$}
Until now the number of classes $Q$ in the unlabelled traffic scenario dataset is assumed to be given. In this study, $Q$ in the unlabelled traffic scenarios is estimated as suggested  in~\cite{Han2019}.  In~\cite{Han2019}, the number of clusters in the unlabelled samples are found using the Silhouette index~\cite{Rousseeuw1987}. The experiment setting is identical to the one used for clustering with $4$ labelled and $3$ unlabelled data. The $Q$ is estimated to be $4$ compared to the ground truth (GT) 3 as seen in the Table~\ref{Tab:highDNo}.
\begin{table}[h]
    \vspace{2mm} 
	\centering
	\caption{Number of classes.}
	\begin{tabular}{| c | c | c | c |} 
		\hline
		Method &  GT  \# of classes & Predicted  \# of classes  & Error \\ 
		\hline\hline
		RFAPs & $3$ & $4$ & $1$\\  
		\hline
	\end{tabular}
	\label{Tab:highDNo}
\end{table}

\subsection{Randomly Chosen Labelled Classes}

This study is conducted to analyse the influence of the chosen labelled and unlabelled classes. Until now, out of the $7$ scenarios, the first $4$ were considered labelled and the remaining $3$ were considered unlabelled. Here, three possible combinations of  labelled and unlabelled classes are chosen randomly  and the clustering performance is analysed.  As shown in  Table~\ref{Tab:highDRatio}, the results remain consistent.

\begin{table}[h!]
	\centering
	\caption{ACC for randomly chosen unlabelled classes.}
	\begin{tabular}{| c | c | c |} 
		\hline
		Labelled classes & Unlabelled classes &   ACC ($\uparrow$)  \\ 
		\hline\hline
		1,2,3,4 & 5,6,7 & 0.810 \\  
		2,3,6,7 & 5,1,4 & 0.82 \\   
		1,2,4,5 & 3,6,7 & 0.807 \\   
		\hline
	\end{tabular}
	\label{Tab:highDRatio}
\end{table}
\subsection{Importance of Step II}
The aim in this study is to understand the influence of using the labelled data in the clustering step (step III) of the method. To perform this, step II is skipped in the proposed method where 3D CNN is fine tuned with labelled data. In the step III, the labelled data is not used and the categorical cross-entropy loss is also dropped. The experiment set-up has $4$ labelled classes and the remaining $3$ classes are considered as unlabelled data. The clustering accuracy with and without the labelled data is shown in Table.~\ref{Tab:highDb}. The results shows experimentally that the labelled data indeed helps in guiding the clustering process. 
\begin{table}[h]
	\centering
	\caption{ACC with and without labelled data.}
	\begin{tabular}{| c | c |} 
		\hline
		Method &  ACC ($\uparrow$)  \\ 
		\hline\hline
		with labelled data & \textbf{0.810} \\  
		without labelled data & 0.565\\
		\hline
	\end{tabular}
	\label{Tab:highDb}
\end{table}
\subsection{Importance of Step III}

\begin{figure}[h!]
	\centering
	\begin{subfigure}{0.45\columnwidth}
		\includegraphics[width=\columnwidth]{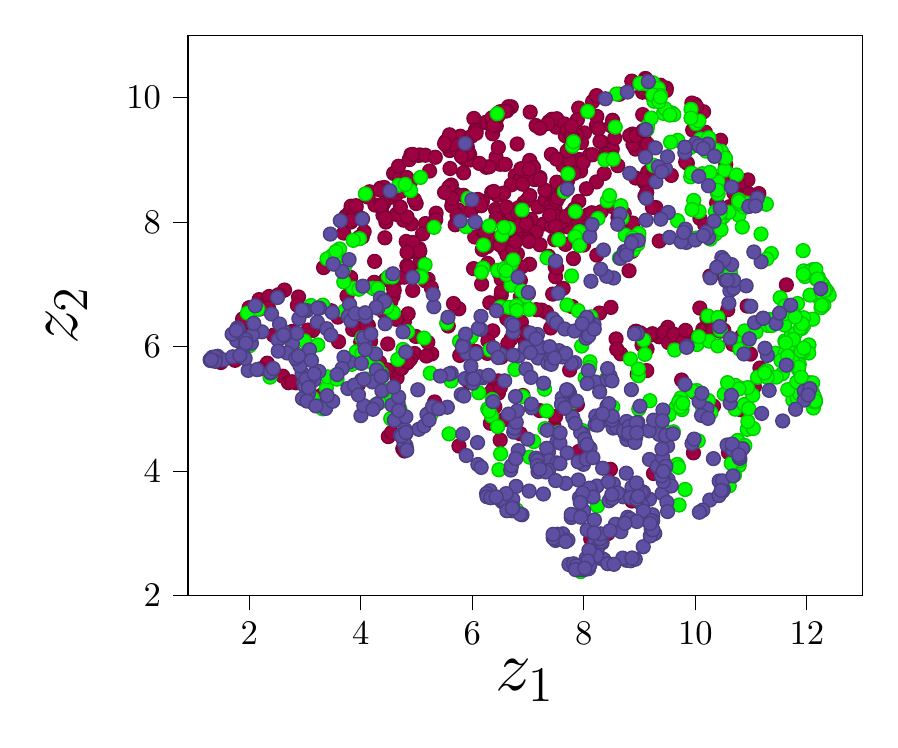}
		\caption{Before optimisation}
		\label{Fig:StepII}
	\end{subfigure}
	\begin{subfigure}{0.45\columnwidth}
		\includegraphics[width=\columnwidth]{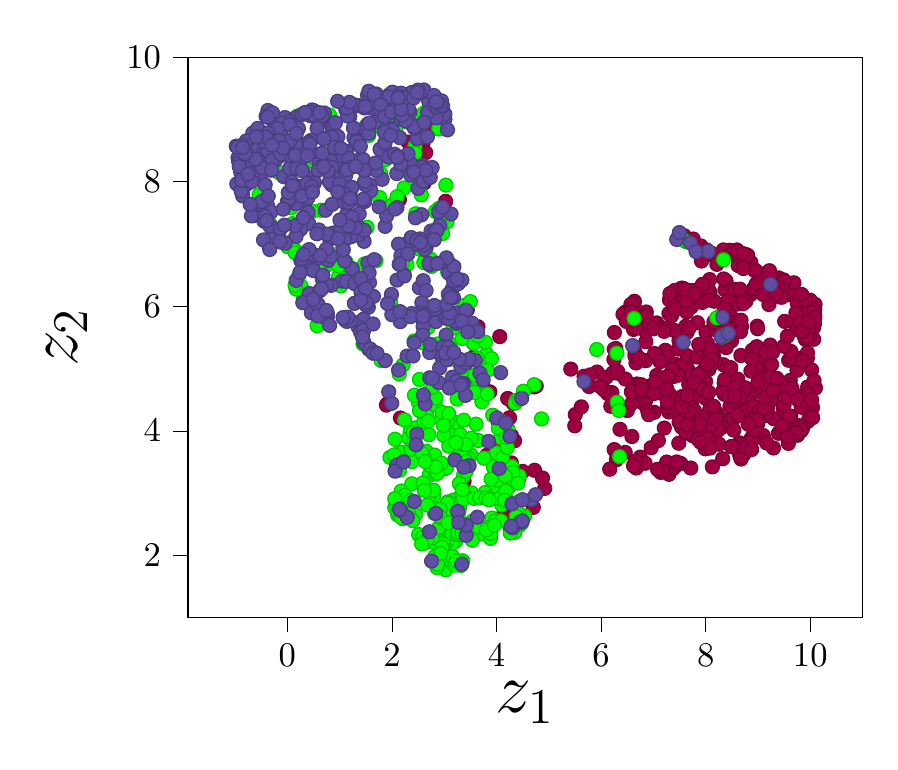}
		\caption{After optimisation}
		\label{Fig:StepIII}
	\end{subfigure}
	\caption{Feature representation study,~\legendsquare{fill=green!70}~Ego following,~\legendsquare{fill=purple!105}~Cut-in from right,~\legendsquare{fill=blue!75}~Leader cut-out to right.}
	\label{Fig:highDRep} 
\end{figure}

The intent of the iterative optimisation in Section~\ref{Sec:Meth} is to provide good feature representations for the unlabelled samples. The experiment setup is similar to the one used in Section~\ref{Sec:Exp}. The feature representations for the considered $3$ unlabelled classes are extracted. The feature representations  before and after the iterative optimisation are projected on to a $2$D space by UMAP~\cite{McInnes2019} as shown in the Fig.~\ref{Fig:highDRep}. It can be seen that there is a good separation of clusters. The clusters, the leader cut-out from ego lane to the right lane (class $7$) and the ego following a leader (class $6$) are close and few datapoints are mixed up. This is because in both of these scenarios there is a leader vehicle in front of the ego in most of the time for a scenario. The clustering accuracy of the $3$ classes before iterative optimisation is $48.65$\% and after iterative optimisation is  $81.0$\%. This shows that the iterative optimisation procedure improves the clustering accuracy.

\section{Conclusion}
\label{Sec:Con}
Traffic scenario clustering is an important problem to be solved for identifying relevant and new traffic scenarios. The new traffic scenario categories are important  for the development of motion planning algorithms and the validation of autonomous functionalities. This work proposes a method to cluster traffic scenarios automatically without any handcrafted features using self-supervised learning and a data-adaptive similarity based on novel features called RFAPs. The problem set-up in this work uses labelled scenarios and retains the knowledge about labelled scenarios to aid the clustering task.  The 3D CNN is trained robust feature representation on a defined pretext task followed by fine-tuning using labelled traffic scenarios. The clustering is addressed by an iterative optimisation procedure using the labelled and unlabelled traffic scenarios. Experiments on the highD dataset  have verified the advantages of the solution proposed when compared to considered baseline methods.


\section*{Acknowledgement}

This work is supported by Bavarian State Ministry for Science and Art under the funding code VIII.2-F1116.IN/18/2.


\bibliographystyle{IEEEtran}
\bibliography{format,reference_new}

\end{document}